\pdfoutput=1
\documentclass[hidelinks,onefignum,onetabnum]{siamart250211}

\usepackage{booktabs}
\usepackage{tikz}
\usepackage{subfig}
\newcommand{\bfx}{\mathbf{x}}

\newcommand{\bfp}{\mathbf{p}}
\newcommand{\bfq}{\mathbf{q}}
\newcommand{\bfW}{\mathbf{W}}
\newcommand{\bftheta}{\boldsymbol{\theta}}
\newcommand{\bfA}{\mathbf{A}}
\newcommand{\bfI}{\mathbf{I}}
\newcommand{\bfy}{\mathbf{y}}
\newcommand{\bfY}{\mathbf{Y}}
\usepackage{enumitem}
\usepackage{multirow}
\usepackage{longtable}

\usepackage{array}
\usepackage{ragged2e}
\newcolumntype{P}[1]{>{\raggedright\arraybackslash}p{#1}}

\usepackage{lipsum}
\usepackage{amsfonts}
\usepackage{graphicx}
\usepackage{epstopdf}
\usepackage{algorithmic}
\ifpdf
  \DeclareGraphicsExtensions{.eps,.pdf,.png,.jpg}
\else
  \DeclareGraphicsExtensions{.eps}
\fi

\newsiamremark{remark}{Remark}
\newsiamremark{hypothesis}{Hypothesis}
\crefname{hypothesis}{Hypothesis}{Hypotheses}
\newsiamthm{claim}{Claim}
\newsiamremark{fact}{Fact}
\crefname{fact}{Fact}{Facts}

\headers{Reversing LLMs for Efficient Training and Fine-Tuning}{E. Gal, M. Eliasof, J. Turek, U. Ascher, E. Treister, and E. Haber}

\title{Reversing Large Language Models for Efficient Training and Fine-Tuning}

\author{
Eshed Gal\thanks{Department of Computer Science, University of British Columbia 
  (\email{eshedg@cs.ubc.ca}).}
\and
Moshe Eliasof\thanks{Department of Computer Science, Ben Gurion University 
  (\email{eliasof@post.bgu.ac.il}).}
\and
Javier Turek\thanks{EarthDynamics AI 
  (\email{javier@earthdynamics.ai}).}
\and
Uri Ascher\thanks{Department of Computer Science, University of British Columbia 
  (\email{ascher@cs.ubc.ca}).}
\and
Eran Treister\thanks{Department of Computer Science, Ben Gurion University 
  (\email{erant@bgu.ac.il}).}
\and
Eldad Haber\thanks{Department of Earth, Ocean and Atmospheric Sciences, University of British Columbia 
  (\email{ehaber@eoas.ubc.ca}).}
}
\usepackage{amsopn}

\ifpdf
\hypersetup{
  pdftitle={Reversing Large Language Models for Efficient Training and Fine-Tuning},
  pdfauthor={D. Doe, P. T. Frank, and J. E. Smith}
}
\fi

\begin{document}

\maketitle

% REQUIRED
\begin{abstract}
Large Language Models (LLMs) are known for their expensive and time-consuming training. Thus, oftentimes, LLMs are fine-tuned to address a specific task, given the pretrained weights of a pre-trained LLM considered a foundation model. In this work, we introduce memory-efficient, reversible architectures for LLMs, inspired by symmetric and symplectic differential equations, and investigate their theoretical properties. Different from standard, baseline architectures that store all intermediate activations, the proposed models use time-reversible dynamics to retrieve hidden states during backpropagation, relieving the need to store activations. This property allows for a drastic reduction in memory consumption, allowing for the processing of larger batch sizes for the same available memory, thereby offering improved throughput. In addition, we propose an efficient method for converting existing, non-reversible LLMs into reversible architectures through fine-tuning, rendering our approach practical for exploiting existing pre-trained models. Our results show comparable or improved performance on several datasets and benchmarks, on several LLMs, building a scalable and efficient path towards reducing the memory and computational costs associated with both training from scratch and fine-tuning of LLMs.
\end{abstract}

% REQUIRED

% REQUIRED

\section{Introduction}

Large Language Models (LLMs) have achieved remarkable success in natural language processing, powering breakthroughs across diverse applications \cite{brown2020language, touvron2023llama}. However, their rapid scaling has introduced critical computational bottlenecks—chief among them is the excessive {memory consumption} during training and fine-tuning.

That limitation is common to most Transformer-based architectures, that require storing intermediate activations, to enable gradient computation via backpropagation. As a result, memory usage grows linearly with depth, significantly constraining the training of deep models and their deployment and training on resource-limited hardware. This limitation not only restricts model size, but also forces smaller batch sizes, which increases training time by requiring more iterations and communication to process the same dataset.

To address this limitation, we draw inspiration from time-reversible physical systems such as wave propagation, advection, and Hamiltonian flows \cite{ruthotto2019deep, haber2017stable, eliasof2020diffgcn}. In particular, we generalize and extend recent advances in reversible vision transformers \cite{Zhao_2024_CVPR, mangalam2022reversible} to the language modeling domain. We alleviate the memory limitation by introducing a family of \emph{reversible architectures} for LLMs based on hyperbolic differential equations, implemented through learned reversible dynamics.

% \paragraph{Our Solution: Reversible Hyperbolic Dynamics.}
% To address the memory limitations, we propose a family of \emph{reversible architectures} for LLMs based on hyperbolic differential equations, implemented through learned reversible dynamics. Drawing inspiration from time-reversible physical systems such as wave propagation, advection, and Hamiltonian flows \cite{ruthotto2019deep, haber2017stable, eliasof2020diffgcn}, we generalize and extend recent advances in reversible vision transformers \cite{Zhao_2024_CVPR, mangalam2022reversible} to the language modeling domain.

A core advantage of our approach is that our architectures are \emph{reversible by design}, allowing the backward pass to reconstruct intermediate activations without storing them during the forward pass. This drastically reduces memory requirements by an order of magnitude, enabling larger batch sizes and thereby significantly improving training throughput. Although reversible methods incur a modest increase in floating-point operations, we show that due to the increase in throughput, the memory savings compensate for the additional computations, reducing wall-clock training time under fixed computational budgets. In summary, training a reversible network results in a reduction in computational time. This idea is similar to ``Activation Recomputation'' \cite{KorthikantiCLMA23}, the difference is that our reversible network does not require storing \emph{any activations}.

%\paragraph{Energy Preservation and Depth Scalability.}
Another distinctive property of our architectures, which are based on hyperbolic PDEs, is their ability to \emph{conserve energy over time}, in contrast to parabolic systems such as diffusion-based networks, which dissipate energy as depth increases \cite{chang2019antisymmetric, gravina2022anti, gravina2025oversquashing}. By modeling token dynamics as conservative flows, our architectures preserve the fidelity of representations across depth, making them more scalable. This inductive bias supports stable long-range information propagation and enables the conceptual design of models with \emph{infinite effective depth}.

%Lastly,  while reversible architectures offer multiple advantages, as discussed above, it is still the case that in many scenarios, it is desired to fine-tune a given large non-reversible model to a specific task using modest resources. To this end, we introduce a novel method for \emph{retrofitting pre-trained non-reversible models into reversible ones}. By leveraging a structural correspondence between reversible dynamics and standard residual updates, we demonstrate that only minimal fine-tuning is needed to enable reversible execution—offering a practical path toward memory-efficient fine-tuning in constrained settings.
Lastly, although reversible architectures offer several advantages, there remain many scenarios where fine-tuning a large non-reversible model for a specific task using limited resources is desirable. To address this, we propose a novel method for \emph{retrofitting pre-trained non-reversible models into reversible ones}. By leveraging a structural correspondence between reversible dynamics and standard residual updates, we show that only minimal fine-tuning is required to enable reversible execution. This provides a practical approach to memory-efficient fine-tuning under resource constraints.

\textbf{Our contributions} are summarized as follows:
\begin{itemize}
    % \item We propose a new class of \textit{reversible language model architectures} derived from hyperbolic differential equations and study their \textit{theoretical properties}.
    \item We propose a new class of \textit{reversible language model architectures} derived from hyperbolic differential equations, leveraging \textit{conservative discretizations} that preserve information flow and guarantee reversibility. We further study their \textit{theoretical properties}.
    \item We demonstrate that these models are \textit{invertible by construction}, enabling memory-efficient training that supports larger batch sizes and enhances \allowbreak\ throughput.
    \item We present a mechanism for \textit{retrofitting pre-trained non-reversible baseline models} into reversible architectures via fine-tuning compatible with existing LLMs.
    \item We present an empirical evaluation of our reversible models, including both training from scratch and retrofitting baseline models into reversible architectures. Our results demonstrate competitive or improved performance, highlighting not only the computational efficiency of reversible LLMs but also their effectiveness on downstream tasks.
\end{itemize}

%%%%%%%%

\section{Reversible Large Language Models}
\label{sec:method}
In this section, we introduce our reversible LLMs. We start with an overview of existing, non-reversible architectures, followed by a formal definition of reversible LLMs. We then discuss the memory footprint of our models, followed by a discussion of their expressiveness and downstream performance compared with standard architectures.
\paragraph{Common Network Architecture Overview}
Modern language models are typically structured in three main stages: (i) an input embedding layer; (ii) a deep core of repeated Transformer layers; (iii) a final projection layer.

Given a tokenized input sequence $\bfx = (\bfx_1, \bfx_2, \ldots, \bfx_T)$, $\bfx_i \in \mathbb{N}$, the model first maps each token to a high-dimensional embedding space $\mathbb{R}^d$, via a learned or pre-defined embedding layer $E: \mathcal{V}\rightarrow \mathbb{R}^{d}$, where $\mathcal{V}$ is the vocabulary index set. This step yields an embedded sequence:
\begin{eqnarray}
    \label{eq:emb}
    \bfp^{(0)} = E(\bfx) + P,
\end{eqnarray}
where $P \in \mathbb{R}^{T \times d}$ represents positional encodings added to the token embeddings. The embedded sequence $\bfp^{(0)} \in \mathbb{R}^{T \times d}$ is then processed by a stack of $L$ layers composed of multi-head self-attention and feedforward multilayer-perceptrons (MLPs), organized in a residual architecture.

Each layer $\ell=1,\ldots,L$ updates the hidden representation via a two-step transformation:
\begin{eqnarray}\label{eq:block}
\bfq^{(\ell)} &=& \bfp^{(\ell-1)} + \mathsf{Attn}_\ell\left( \mathsf{LN}_1(\bfp^{(\ell-1)}) \right) \nonumber\\
\bfp^{(\ell)} &=& \bfq^{(\ell)} + \mathsf{MLP}_\ell\left( \mathsf{LN}_2(\bfq^{(\ell)}) \right),
\end{eqnarray}
where $\mathsf{Attn}_\ell$ is the multi-head self-attention module, $\mathsf{MLP}_\ell$ is a feedforward network, and $\mathsf{LN}_1, \mathsf{LN}_2$ denote layer normalization applied before each sub-block. This structure forms a residual update scheme in which $\bfp^{(\ell)}$ is incrementally refined across layers by injecting nonlinear attention updates, while maintaining a residual path for stable gradient flow.

After $L$ such updates, the final representation $\bfp^{(L)}$ is mapped back to the vocabulary logits using a final projection layer that reads: 
\begin{eqnarray}
    \label{eq:closing}
\hat{\bfx} = \mathrm{softmax}\left( \bfW \bfp^{(L)}  \right),
\end{eqnarray}
where $\bfW$ are learned weights. 
This step yields a probability distribution over next tokens for autoregressive generation or language modeling objectives.
While effective, this architecture demands storing all intermediate representations
$(\bfp^{(0)}, \bfq^{(0)}), \ldots, \allowbreak\ (\bfp^{(L)}, \bfq^{(L)})$ for backpropagation, and lacks a principled constraint on how representational energy is preserved or dissipated through the depth of the network — motivating our proposed reversible, energy-conserving LLMs.

\subsection{Reversible Architecture} 

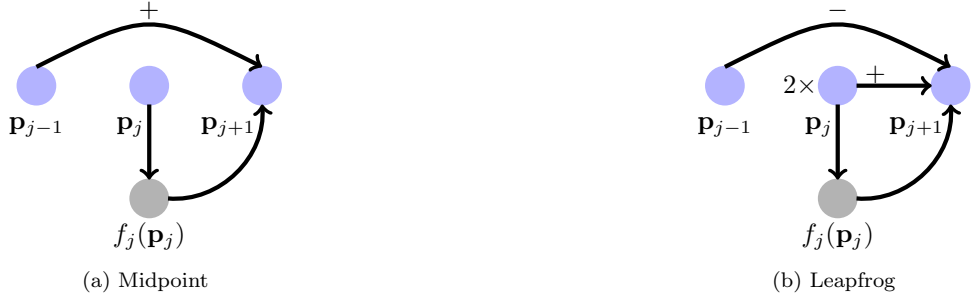
\begin{figure}[t]
\centering

\subfloat[Midpoint\label{fig:arch_midpoint}]{%
\begin{tikzpicture}[thick,scale=1]
    \filldraw[blue!30] (-1.5,0) circle (0.25);
    \node[scale=1] at (-1.5,-0.55) {$\bfp_{j-1}$};
    
    \filldraw[blue!30] (0,0) circle (0.25);
    \node[scale=1] at (-0.25,-0.55) {$\bfp_{j}$};
    
    \filldraw[blue!30] (1.5,0) circle (0.25);
    \node[scale=1] at (1.05,-0.55) {$\bfp_{j+1}$};
    
    \filldraw[gray!60] (0,-1.5) circle (0.25);
    \node[scale=1] at (0,-2.0) {$f_{j}(\bfp_j)$};
    
    \draw[->, ultra thick] (-1.5,0.25) .. controls (0,1) .. (1.5,0.25);
    \node[scale=1] at (0,1) {$+$};
    
    \draw[->, ultra thick] (0,-0.25) -- (0, -1.25);
    \draw[->, ultra thick] (0.25,-1.5) to[bend right=50] (1.5, -0.25);
\end{tikzpicture}
}
\hfill
\subfloat[Leapfrog\label{fig:arch_leapfrog}]{%
\begin{tikzpicture}[thick,scale=1]
    \filldraw[blue!30] (-1.5,0) circle (0.25);
    \node[scale=1] at (-1.5,-0.55) {$\bfp_{j-1}$};
    
    \filldraw[blue!30] (0,0) circle (0.25);
    \node[scale=1] at (-0.25,-0.55) {$\bfp_{j}$};
    
    \filldraw[blue!30] (1.5,0) circle (0.25);
    \node[scale=1] at (1.05,-0.55) {$\bfp_{j+1}$};
    
    \filldraw[gray!60] (0,-1.5) circle (0.25);
    \node[scale=1] at (0,-2.0) {$f_{j}(\bfp_j)$};
    
    \draw[->, ultra thick] (-1.5,0.25) .. controls (0,1) .. (1.5,0.25);
    \node[scale=1.0] at (0,1) {$-$};
    
    \node[scale=1.0] at (0.5,0.15) {$+$};
    \draw[->, ultra thick] (0.25, 0) -- (1.25, 0);
    
    \draw[->, ultra thick] (0,-0.25) -- (0, -1.25);
    \node[scale=1.0] at (-0.5,0) {$2 \times$};
    
    \draw[->, ultra thick] (0.25,-1.5) to[bend right=50] (1.5, -0.25);
\end{tikzpicture}
}

\caption{Reversible architectures: (a) explicit midpoint update and (b) leapfrog update.}
\label{fig:archs}
\end{figure}

%While reversible architectures have been used in computer vision, there is only limited work that uses them in language. In Reformer \cite{kitaev2020reformerefficienttransformer} and PaReprop \cite{zhu2023parepropfastparallelizedreversible}, the authors proposed Hamiltonian-like dynamics, also studied in \citet{haber2017stable}. In this work, we extend the ideas and propose two new reversible architectures that are summarized in Figure~\ref{fig:archs}, and explore conditions for a reversible architecture to be forward as well as backward stable.
While reversible architectures have been explored in computer vision, their application to language modeling remains limited. Reformer \cite{kitaev2020reformerefficienttransformer} and PaReprop \cite{zhu2023parepropfastparallelizedreversible} introduced reversible models with Hamiltonian-like dynamics, building on ideas also studied in \cite{haber2017stable}. In this work, we leverage these foundations and introduce two novel reversible architectures, summarized in Figure~\ref{fig:archs}. We also investigate conditions under which a reversible architecture is both forward and backward stable.

\paragraph{Midpoint Discretization}
We begin by introducing a reversible architecture inspired by an explicit midpoint integrator commonly used for time-reversible numerical integration of first-order differential equations. In this formulation, the hidden state evolution is governed by a discretized update rule of the form:
\begin{align}
\label{eq:midpoint}
\bfp^{(\ell+1)} = \bfp^{(\ell-1)} + 2h  f_{\bftheta_\ell}(\bfp^{(\ell)}),
\end{align}
where $\bfp^{(\ell)} \in \mathbb{R}^{T \times d}$ denotes the hidden state at layer $\ell$, $h > 0$ is a fixed step size, and $f_{\theta_\ell}$ is a learnable update function applied to the current state. This formulation is reversible by construction: given $\bfp^{(\ell+1)}$ and $\bfp^{(\ell)}$, the previous state $\bfp^{(\ell-1)}$ can be exactly recovered via:
$$
\bfp^{(\ell-1)} = \bfp^{(\ell+1)} - 2h \, f_{\theta_\ell}(\bfp^{(\ell)}).
$$

The update function $f_{\theta_\ell}$ encapsulates a full transformer block composed of attention and MLP submodules in \eqref{eq:block}. Specifically, we define:
\begin{eqnarray}
\label{eq:layer}
f_{\theta_\ell}(\bfp) &=& \mathsf{Attn}_\ell\left( \mathsf{LN}_1(\bfp) \right) + \\
\nonumber
&& \mathsf{MLP}_\ell\left( \mathsf{LN}_2\left(\bfp + \mathsf{Attn}_\ell\left( \mathsf{LN}_1(\bfp) \right) \right) \right),
\end{eqnarray}
%This ordering of operations 
which mirrors the standard attention–MLP sequence in transformer blocks. In \eqref{eq:midpoint}, this standard block is wrapped in a midpoint-style update that conserves information across layers. In this architecture, $h$ is chosen as a hyper-parameter. 
% The midpoint method, widely used in physics-based simulations \cite{ascher2008numerical}, is symplectic and time-reversible, making it well-suited to neural architectures where energy preservation and invertibility are desirable. 
Notably, this architecture maintains full reversibility across depth, enabling memory-efficient backpropagation without caching intermediate states, and provides a natural inductive bias for non-decaying representation flow.

\paragraph{Leapfrog Discretization.}
%While the midpoint-based architecture in \Eqref{eq:midpoint} is reversible and efficient, it can exhibit sensitivity to real positive eigenvalues in the Jacobian of $f_{\theta_\ell}$, leading to exponential divergence or instability during training  \cite{ascher2008numerical}. To improve stability while preserving reversibility and long-range propagation, we propose an additional architecture, motivated by second-order hyperbolic PDEs, particularly the form of the nonlinear wave equation. Such architecture was studied in detail in \cite{ruthotto2019deep,eliasof2021pde}, although not for LLMs. In this formulation, the layerwise update is given by:
%\begin{align}
%\label{eq:leapfrog}
%\bfp^{(\ell+1)} = 2\bfp^{(\ell)} - \bfp^{(\ell-1)} + h^2 f_{\theta_\ell}(\bfp^{(\ell)}).
%\end{align}
%This equation corresponds to a finite-difference discretization of the second-order system 
%\begin{equation}\label{eq:second-order}
%\ddot{ \bfp} = f_{\theta}(\bfp),
%\end{equation}
%where time is indexed by discrete layer depth $\ell$, and  $h>0$ is the step size. Like the leapfrog integrator, this update is fully reversible, as previous states can be exactly reconstructed from future ones, by reversing the equation backwards.

While the midpoint-based architecture in \eqref{eq:midpoint} is both reversible and efficient, it can exhibit sensitivity to real positive eigenvalues in the Jacobian of $f_{\theta_\ell}$, potentially leading to exponential divergence or instability during training \cite{ascher2008numerical}. To improve stability while preserving reversibility and long-range information propagation, we propose an alternative architecture inspired by second-order hyperbolic partial differential equations—specifically, the nonlinear wave equation. Such an architecture has been studied in detail in \cite{ruthotto2019deep, eliasof2021pde}, though not in the context of large language models. In this formulation, the layerwise update is given by:
\begin{align}
\label{eq:leapfrog}
\bfp^{(\ell+1)} = 2\bfp^{(\ell)} - \bfp^{(\ell-1)} + h^2 f_{\theta_\ell}(\bfp^{(\ell)}),
\end{align}
which corresponds to a finite-difference discretization of the second-order system:
\begin{equation}\label{eq:second-order}
\ddot{ \bfp} = f_{\theta}(\bfp),
\end{equation}
where time is indexed by the discrete layer depth $\ell$, and $h > 0$ denotes the step size. Like the leapfrog integrator, this update is fully reversible, as previous states can be exactly reconstructed by integrating the update backward.

The second-order formulation in \eqref{eq:leapfrog} yields increased numerical stability and improved handling of oscillatory dynamics, especially when $f_{\theta_\ell}$ has eigenstructure with nonzero real parts. Importantly, the update conserves a discrete analogue of energy over layers, akin to physical systems governed by Hamiltonian or wave dynamics. This makes it well-suited for LLMs tasked with propagating semantic information across long contexts without dissipation.

\paragraph{Hamiltonian Dynamics.}
We now introduce a third reversible architecture motivated by Hamiltonian dynamics,
in which the evolution of the system is governed by coupled first-order differential equations describing the interaction between positions and momenta. A similar system was introduced in \cite{haber2017stable} and recently explored for vision transformers \cite{mangalam2022reversible}. To model this structure, we maintain two coupled hidden states: $\bfp^{(\ell)} \in \mathbb{R}^{T \times d}$, interpreted as the "position", and $\bfq^{(\ell)} \in \mathbb{R}^{T \times d}$, interpreted as the "momentum". The model evolves these states through a staggered update scheme resembling the symplectic Euler integrator for Hamiltonian systems:
\begin{align}
\label{eq:hamil}
\bfq^{(\ell)} &= a_{\ell-1}\bfq^{(\ell-1)} + \mathsf{Attn}_\ell\left( \mathsf{LN}_1(\bfp^{(\ell-1)}) \right) \\
\bfp^{(\ell)} &= b_{\ell-1}\bfp^{(\ell-1)} + \mathsf{MLP}_\ell\left( \mathsf{LN}_2(\bfq^{(\ell)}) \right).
\end{align}
where we choose $a_{\ell}=b_{\ell}=1$ (see next section).
Each update alternates between applying a self-attention update to the position-like variable $\bfq$, and an MLP-based update to the momentum-like variable $\bfp$. This structure mirrors the canonical Hamiltonian equations $\dot{\bfp} = \nabla_{\bfq} H$, $\dot{\bfq} = -\nabla_{\bfp} H$, and conserves a discrete analogue of total energy, as well as flow volume conservation properties. Notably, the system is reversible, as each state can be recovered from its successor using the inverse of the update function.

This staggered formulation not only improves stability compared to direct residual updates, but also introduces a natural division of computational roles: attention governs the flow of global context through momentum-like variables, while MLPs serve as local content-based forces on the positional trajectory. The resulting dynamics support smooth, energy-preserving, volume-preserving propagation of information across depth, further enhancing the model’s capacity to maintain long-range context without diffusion.

\subsection{Memory Footprint}

%Reversible architectures achieve substantial memory savings during training by eliminating the need to store intermediate activations for gradient computation. In conventional networks such as transformers, each layer’s output must be stored during the forward pass to enable gradient computation during the backward pass, resulting in a memory footprint that scales linearly with the number of layers. \textit{In contrast}, our reversible networks reconstruct intermediate activations on-the-fly during backpropagation using their invertible update rules (i.e., \Eqref{eq:midpoint}, \Eqref{eq:leapfrog}, or \Eqref{eq:hamil}), so only the inputs and final outputs need to be retained, yielding a constant activation memory cost, independent of model depth. We illustrate this property in Figure \ref{fig:memory_consumption}.
Reversible architectures enable substantial memory savings during training by eliminating the need to store intermediate activations for gradient computation. In conventional networks such as transformers, each layer’s output must be retained during the forward pass to compute gradients, resulting in a memory footprint that scales linearly with depth. In contrast, our reversible networks reconstruct intermediate activations on-the-fly during backpropagation using invertible update rules (e.g., \eqref{eq:midpoint}, \eqref{eq:leapfrog}, \eqref{eq:hamil}), so only inputs and final outputs must be stored. This yields a constant activation memory cost, independent of model depth. We illustrate this behavior in Figure~\ref{fig:memory_consumption}.

%This memory efficiency comes at a moderate computational cost: during the backward pass, each layer’s forward function must be re-evaluated to reconstruct activations needed for computing gradients. However, this recomputation does not double the total computation time. In fact, because computing the gradient of a layer (e.g., via automatic differentiation) is often significantly more expensive than evaluating the layer itself, the additional cost of redoing the forward function typically adds only 30–50\% overhead. The precise figure depends on the complexity of the layer, for example, MLPs and attention layers with high arithmetic intensity tend to have more costly backward steps, so the marginal cost of re-evaluating $f$ is relatively low.
This memory efficiency comes at a moderate computational cost: during the backward pass, each layer’s forward function must be re-evaluated to reconstruct activations for gradient computation. However, this recomputation does not incur a two-fold increase in compute time. In practice, gradient computation, e.g., via automatic differentiation, is often significantly more expensive than the forward pass, so re-evaluating the layer typically adds only 30–50\% overhead. The exact cost depends on layer complexity; for instance, MLPs and attention modules with high arithmetic intensity tend to have costly backward steps, making the marginal cost of recomputing $f_\theta$ relatively low.

%It is important to note that when using a GPU as a computational hardware, the reduction in memory, {\em even with an increased flop-count}, can reduce the overall compute time. This is because the run time does not scale linearly with the batch size.  This implies that an algorithm that can use a bigger batch size can actually pass more data through the network in a shorter amount of time, even with a slower network. The computational saving varies based on the exact architecture and memory usage. We discuss this further in the numerical experiment section.
Notably, on GPU hardware, reduced memory usage, even with increased FLOPs, can improve overall runtime, as runtime does not scale linearly with batch size. Algorithms that support larger batches may process more data per unit time, even if individual forward passes are slower. The exact performance gain depends on the architecture and memory footprint. We discuss this further in Section~\ref{sec:experiments}.

\subsection{Reversibility Impact on Model Quality}

A natural question arises when introducing structured, reversible dynamics in place of conventional dynamics: does modifying the architecture's internal dynamics degrade the model’s expressivity or performance? While reversible updates constrain the form of information flow, our experiments reveal that these constraints do not hinder model quality, at least on the models that can be run with modest resources. In fact, we find that models trained with our energy-conserving reversible architectures not only match the performance of standard transformer-based language models, but often \emph{outperform them}, as we show in Section \ref{sec:experiments}.

%This performance gain appears to stem from two sources. First, the structured flow of information, governed by time-reversible dynamics, encourages better global propagation of representations and less reliance on shortcut pathways \citep{gravina2025oversquashing}. Second, the conservation of representational energy across layers mitigates the vanishing or diffusion effects common in deep neural network stacks, enabling more stable gradient flow and richer hidden state evolution \citep{chang2019antisymmetric}. As a result, the architecture’s theoretical advantages in memory and stability translate into tangible improvements in generalization and task performance, even when no explicit memory constraints are present. 
This performance gain appears to stem from two sources. First, the structured flow of information, governed by time-reversible dynamics, encourages better global propagation of representations and reduced reliance on shortcut pathways \cite{gravina2025oversquashing}. Second, the conservation of representational energy across layers mitigates the vanishing or diffusion effects common in deep neural network stacks, enabling more stable gradient flow and richer hidden state evolution \cite{chang2019antisymmetric}. Hence, the architecture’s theoretical advantages in memory and stability translate into tangible improvements in generalization and task performance, even when no explicit memory constraints are present.

\section{Analysis of Reversible Methods}
In this Section, we analyze the stability of reversible network architectures, beginning with constant-coefficient formulations and extending to the practical case of variable coefficients encountered in deep learning.
\subsection{Reversibility for Constant Coefficients}
While reversible methods have been proposed for deep networks, it is important to note that reversible networks are typically only \textit{marginally stable}, and this can lead to practical difficulties when attempting to train them. 
To this end, we first study a reversible method of the form
\begin{eqnarray}
    \label{revnetfor}
    \bfp_{j+1} = a \bfp_{j-1} + b \bfp_j + h f(\bfp_j).
\end{eqnarray}
As commonly done to estimate stability \cite{apbook}, we study the test equation $f(\bfp) = \lambda \bfp$, where $\lambda$ represents the Jacobian of the nonlinear transformation.

This type of equation is solved by assuming a solution of the form $\bfp_j = r^j$. Substituting in  \eqref{revnetfor} we obtain:
\begin{equation}
    r^2 - (b+h\lambda) r - a = 0.
\end{equation}
This is a quadratic equation with two, possibly complex, roots. For the stability of the forward pass, we need both solutions to have absolute values equal at most $1$.
However, we look for stability of both forward and backward processes. To this end we seek solutions where $|r| = 1$, which implies that the solution is:
\begin{equation}
    r_{12} = \exp(\pm i \theta).
\end{equation}
Because 
$r_1 r_2 = a$ we obtain that
$|a|=1$ is a necessary condition for stability. 

Substituting the solutions and $a = \pm 1$ in the equation  and dividing by $\exp(i \theta)$ we obtain: 
\begin{eqnarray}
   \exp(i \theta) \mp \exp(-i \theta) = \left\{\begin{matrix} 2 i\sin(\theta) \\ 2\cos(\theta) \end{matrix} \right.   = b + \lambda h .
\end{eqnarray}
This equation has two main implications. First, it yields a condition on $b$, namely:
\begin{eqnarray}
    \label{stabcond}
    |b + \lambda h| \le 2.
\end{eqnarray}
Second, it yields a condition on $\lambda$. Note that,  if $\lambda$ is complex, then there is no solution to the system, implying that there is no forward-backward stable method for a complex $\lambda$. Furthermore, for $a=1$, $\lambda$ needs to be purely imaginary while for $a=-1$, $
\lambda$ is requires to be purely real and negative. 
That is, the eigenvalues of the Jacobian should be either purely real (for $a=-1$) or purely imaginary (for $a=1$) to obtain stability. 

%The Hamiltonian-type network in \Eqref{eq:hamil} was also considered as a reversible network. Similar analysis for that dynamics is presented in the Appendix, and it reveals that similarly, the product of coefficients $a_{\ell-1}$ and $b_{\ell-1}$ must be $1$ for forward and backward stability. In this work, we do not use the Hamiltonian dynamics but rather mainly use the midpoint and experiment with the leapfrog network, but we present it for \emph{completeness}. Note that both the Hamiltonian and leapfrog represent {\em second-order dynamics} (i.e., second-order time derivative as in \Eqref{eq:second-order}), while the standard skip connection and the midpoint method represent first-order dynamics (i.e., first-order time derivative). Such dynamical systems have very different properties. While first-order dynamics can be seen as a system for the velocity of the data, second-order systems represent the acceleration. While many networks, such as ResNets \citep{he2016deep}, have demonstrated that first-order techniques can be highly competitive, less research has been conducted for second-order systems.
The Hamiltonian-type network in \eqref{eq:hamil} was also considered as a reversible architecture. A similar analysis, presented below, shows that the product of coefficients $a_{\ell-1}$ and $b_{\ell-1}$ must equal 1 for forward and backward stability. While we focus on the midpoint method and experiment with the leapfrog network, we include the Hamiltonian formulation for \emph{completeness}. Both the Hamiltonian and leapfrog updates represent {\em second-order dynamics} (i.e., involving second-order time derivatives as in \eqref{eq:second-order}), whereas the standard skip connection and midpoint correspond to first-order dynamics. These dynamical systems differ significantly: first-order models describe the velocity of data, while second-order models represent its acceleration. While first-order approaches, such as ResNets \cite{he2016deep}, have proven highly effective, second-order systems remain less explored.

\subsection{Reversibility for Variable Coefficients}

% \jt{The motivation of this subsection is unclear. Why are the coefficients non-constant? Can we show this with an empirical analysis?}
In the previous subsection, we explored the reversibility of a constant coefficient method. While such an analysis is common for the analysis of dynamical systems, deep networks may not be fully conservative, indeed, the network may demand that some modes grow moderately while others shrink. We therefore propose estimating the behavior of a network of the form:
\begin{eqnarray}
    \label{revnetforS}
    \bfp_{j+1} = a_{j-1} \bfp_{j-1} + (1-a_{j-1}) \bfp_j + h f_j(\bfp_j).
\end{eqnarray}
We call this method the midpoint ($a$) method, similar to the fractional $\theta$ methods commonly introduced \cite{ascher2008numerical}.
Assume that we choose 
\begin{eqnarray}
\label{eq:aj}
a_j \sim [1 + {\frac 12}U(-1,1), -1 + {\frac 12}U(-1,1)].
\end{eqnarray}

%\begin{equation}
%\label{eq:aj}
%a_j = \text{Bernoulli}(\pm 1) \cdot \frac{1}{\sqrt{2}} + U(-\sqrt{\frac{3}{2}}, \sqrt{\frac{3}{2}})
%\end{equation}
In this case, we cannot consider the stability of a layer in a deterministic form, but rather in expectation.
To analyze stability in expectation, we take the expectation and the variance of the recurrence relation.
The expectation yields
\begin{eqnarray}
    \label{revnetforSE}
    {\mathbb E}\bfp_{j+1} = {\mathbb E} (a_{j-1} \bfp_{j-1}) + {\mathbb E}((1-a_{j-1}) \bfp_j) + h {\mathbb E} (\lambda_j\bfp_j).
\end{eqnarray}
Note that due to the linearity of the expectation and the choice of distribution of $a_{j-1}$ having a zero mean, the first term vanishes
and we obtain that 
\begin{eqnarray}
    \label{revnetforSE1}
    {\mathbb E}\bfp_{j+1} = {\mathbb E} (a_{j-1} \bfp_{j}) + (1 + h \lambda_j){\mathbb E} \bfp_j = (1 + h \lambda_j){\mathbb E} \bfp_j.
\end{eqnarray}
To compute the variance of a single step, we rearrange the recurrence relation to group terms involving $a_{j-1}$:
\begin{align*}
\bfp_{j+1} = a_{j-1} (\bfp_{j-1} - \bfp_j) +  (1 + h \lambda_j)\bfp_j.
\end{align*}
For the choice of $a_{j-1}$ in \eqref{eq:aj}, we have that
${\mathbb Var}(a_{j-1}) =  1$,
which implies that 
\begin{eqnarray}
\label{eq:var}
{\mathbb Var}(\bfp_{j+1}) = (\bfp_{j-1} - \bfp_j)^2. \end{eqnarray}

Note that this {\em reversible} method behaves in expectation just like the forward Euler equation in terms of stability. As we see in Section \ref{sec:conversion}, this can be used to convert a non-reversible method to a reversible one.

\subsection{Stability of reversible Hamiltonian methods}

A more \allowbreak\ general reversible method is a Hamiltonian-style method of the form
\begin{subequations}
\begin{eqnarray}
    \label{ham}
    \bfp_{j+1} &=& a \bfp_j + f(\bfq_{j}) \\
    \bfq_{j+1} &=& b \bfq_j + g(\bfp_{j+1})
\end{eqnarray}
\end{subequations}

To understand the stability property we study the linear case
\begin{subequations}
\begin{eqnarray}
    \label{ham2}
    \bfp_{j+1} &=& a \bfp_j + \alpha \bfq_j \\
    \bfq_{j+1} &=& b \bfq_j + \beta \bfp_{j+1}  
\end{eqnarray}
\end{subequations}
Rearranging we have that
\begin{eqnarray}
    \label{ham1}
    \bfp_{j+1} &=& a \bfp_j + \alpha \bfq_j \\
    \nonumber
    \bfq_{j+1} &=& b \bfq_j + \beta (a \bfp_j + \alpha \bfq_j)  = a \beta \bfp_j + (b + \alpha \beta) \bfq_j
\end{eqnarray}
or
\begin{eqnarray}
    \label{hammat}
    \begin{pmatrix}
    \bfp_{j+1} \\  \bfq_{j+1}  
    \end{pmatrix} = 
    \begin{pmatrix}
    a & \alpha \\  a \beta & b + \alpha \beta   
    \end{pmatrix} 
    \begin{pmatrix}
    \bfp_{j} \\  \bfq_{j}  
    \end{pmatrix} 
\end{eqnarray}
For forward-backward stability we need that $|\lambda_1| = |\lambda_2| = 1$ which implies that
\begin{eqnarray}
 && |{\rm det}(\bfA)| = |\lambda_1 \lambda_2|  = |ab| = 1  \\
 && {\rm trace}(\bfA) = |a+b+\alpha \beta| \le 2
\end{eqnarray}

\section{From Baseline to Reversible LLMs}
\label{sec:conversion}

While reversible models can be trained from scratch, it is often advantageous to leverage the weights of pre-trained models and convert them into reversible networks. In this section, we present a method to retrofit standard residual architectures into approximately reversible ones, which can optionally be refined through fine-tuning. We further analyze this approximation by introducing  higher-order estimators and studying their accuracy.

\subsection{Approximating Reversibility}
 Consider a residual network of the form:
\begin{eqnarray}
    \label{eq:fe}
    \bfp_{j+1} = \bfp_j + f_j(\bfp_j),
\end{eqnarray}
where $f_j$ denotes the $j$th layer with its own parameters. The update for the previous layer is:
\begin{eqnarray}
    \label{eq:feold}
    \bfp_j = \bfp_{j-1} + f_{j-1}(\bfp_{j-1}).
\end{eqnarray}
Multiplying \eqref{eq:feold} by $a_{j-1}$ and substituting $a_{j-1} \bfp_j$ into \eqref{eq:fe}, we obtain:
\begin{eqnarray}
    \label{eq:almostmid}
    \bfp_{j+1} &=& a_{j-1} \bfp_{j-1} + (1 - a_{j-1}) \bfp_j + \\
    \nonumber
    && \left( f_j(\bfp_j) + a_{j-1} f_{j-1}(\bfp_{j-1}) \right).
\end{eqnarray}
This update recovers the residual structure in \eqref{eq:fe} but resembles the reversible form in \eqref{revnetforS}. However, since the right-hand side depends on both $\bfp_j$ and $\bfp_{j-1}$, the network remains non-reversible.

To restore reversibility, we approximate $\bfp_{j-1}$ using a backward estimate from \eqref{eq:feold}:
\begin{eqnarray}
    \label{eq:pold_est}
    \bfp_{j-1} = \bfp_j - f_{j-1}(\bfp_{j-1}) \approx \bfp_j - f_{j-1}(\bfp_j).
\end{eqnarray}
Substituting this estimate into \eqref{eq:almostmid} yields the reversible update:
\begin{eqnarray}
    \label{eq:almostmid_rev}
    \widehat{\bfp}_{j-1} &=& \bfp_j - f_{j-1}(\bfp_j), \\
    \nonumber
    \bfp_{j+1} &=& a_{j-1} \widehat{\bfp}_{j-1} + (1 - a_{j-1}) \bfp_j + \\
    \nonumber
    && \left( f_j(\bfp_j) + a_{j-1} f_{j-1}(\widehat{\bfp}_{j-1}) \right).
\end{eqnarray}
This approximation yields a fully reversible update that mimics the dynamics of the original residual network. A detailed analysis of the approximation error is provided in the appendix.

While \eqref{eq:almostmid_rev} can be used as a zero-shot approximation, it may also be refined via fine-tuning. Let $\bfY_{\rm fe} = [\bfy_1, \ldots, \bfy_N]$ denote the output probabilities from the original residual model in \eqref{eq:fe}, and let $\bfY_{\rm rev}$ be those from the reversible formulation in \eqref{eq:almostmid_rev}. We fine-tune the model by solving:
\begin{eqnarray}
    \label{min_opt1}
    \min\ {\mathbb E} \left[
    KL\left(\bfY_{\rm rev}, \bfY_{\rm fe}\right)
    \right],
\end{eqnarray}
where the optimization is over the parameters generating $\bfY_{\rm rev}$, while keeping those of $\bfY_{\rm fe}$ fixed.

\subsection{Higher Order Estimators of $\widehat \bfp_{j-1}$ and their analysis}

The quality of the approximation of our network in Equation~\eqref{eq:almostmid_rev} to the original network in Equation~\eqref{eq:fe} relies on the quality
of the approximation to $\bfp_{j-1}$.
In the above, we have used a single-step approximation. A more general idea is to use a fixed-point iteration of the form
\begin{eqnarray}
    \label{eq:fixedpoint}
    \bfp_{j-1}^{(k+1)}  = \bfp_j - f_{j-1}(\bfp_{j-1}^{(k)}) 
\end{eqnarray}
initializing $\bfp_{j-1}^{(0)} = \bfp_{j}$.
The first iteration yields our previous update; however, using more iterations, we obtain a more accurate estimate of the true $\bfp_{j-1}$, which implies a better reversible approximation of the original system.

% \subsection{Analysis}
\subsubsection{Linear analysis}
Above, we have discussed a methodology that approximates a non-reversible architecture with a reversible one. To understand the approximation we analyze the simple linear case.
We consider a network of the form
\begin{eqnarray}
    \label{eq:linear}
    \bfp_{j+1} = 
    a_{j-1} \bfp_{j-1} + (1-a_{j-1})\bfp_j + \bfA_j \bfp_j + a_j \bfA_{j-1} \bfp_{j-1} 
    \end{eqnarray}
This network corresponds to the standard single recurrence network.
The corresponding approximate  architecture reads
\begin{eqnarray}
    \label{eq:approxmid}
     \widehat\bfp_{j+1} &=& a_{j-1} \bfp_{j-1} + (1-a_{j-1})\bfp_j + \bfA_j \bfp_j + \\
     \nonumber
     && a_{j-1} \bfA_{j-1}
     (\bfp_j - \bfA_{j-1}\bfp_{j}) 
\end{eqnarray}
Subtracting \eqref{eq:linear} from \eqref{eq:approxmid} we obtain that
\begin{eqnarray}
    \nonumber
\widehat\bfp_{j+1} - \bfp_{j+1} &=&  
 \bfA_{j-1} \bfp_{j-1} - a_{j-1}(\bfA_{j-1}  - \bfA_{j-1}^2) \bfp_j   \\
 \nonumber
&=& \bfA_{j-1} \bfp_{j-1} - a_{j-1} (\bfA_{j-1}  - \bfA_{j-1}^2) (\bfI + \bfA_{j-1})\bfp_{j-1} \\
\label{diff}
&=&\bfA_{j-1} \left( \bfI  - a_{j-1} (\bfI - \bfA_{j-1}^2 ) \right)\bfp_{j-1}
\end{eqnarray}
Thus we obtain that
\begin{eqnarray}
    \label{eq:error}
&&  \|\widehat\bfp_{j+1} - \bfp_{j+1} \|^2 \le \\
  \nonumber   
&&  \|\bfA_{j-1}\|^2 \| (1-a_{j-1})\bfI - a_{j-1}\bfA_{j-1}^2\|^2 \|\bfp_{j-1}\|^2
\end{eqnarray}
This approximation can be very good when the norm of $\bfA_{j-1}$ is small. The approximation falls apart when $\|\bfA_{j-1} \| > 1$.

Residual networks often exhibit a behavior 
where the norm of the update is small and therefore, our proposed approximation works well for many networks in practice.

\section{Numerical Experiments}
\label{sec:experiments}

\begin{table*}[t]
\centering
    \centering
    \begin{tabular}{lcccc} 
    \toprule
    \textbf{Network} & \multicolumn{2}{c}{\textbf{GPT Small (124M)}} & \multicolumn{2}{c}{\textbf{GPT Large (772M)}} \\
    \cmidrule(lr){2-3} \cmidrule(lr){4-5}
     & \textbf{Training} & \textbf{Validation} & \textbf{Training} & \textbf{Validation} \\
    \midrule
    Baseline & 2.8810 & 2.9022 & 2.6544 & 2.6988 \\
    Midpoint & 2.8681 & 2.9148 & 2.5243 & 2.6183 \\
    Leapfrog & 2.9245 & 2.9432 & 2.5750 & 2.6091 \\
    \bottomrule
    \end{tabular}
    \caption{Training and validation cross-entropy loss($\downarrow$) of baseline and our reversible Midpoint and Leapfrog architectures.}
    \label{tab:gpt2_train_val}
\end{table*}

% \begin{table*}[t]
% \centering
% \begin{tabular}{l ccc ccc}
% \toprule
% & \multicolumn{3}{c}{\textbf{GPT-2 Small (124M)}} & \multicolumn{3}{c}{\textbf{GPT-2 Large (772M)}} \\
% \cmidrule(lr){2-4} \cmidrule(lr){5-7}
% \textbf{Benchmark} & \textbf{Baseline} & \textbf{Midpoint} & \textbf{Leapfrog} & \textbf{Baseline} & \textbf{Midpoint} & \textbf{Leapfrog} \\
% \midrule
% piqa & 51.14\% & 53.05\% & 52.83\% & 53.16\% & 52.72\% & 55.11\% \\
% ARC-e & 27.02\% & 26.14\% & 28.95\% & 23.86\% & 26.84\% & 26.49\% \\
% ARC-c & 22.74\% & 22.73\% & 24.08\% & 22.41\% & 28.09\% & 26.09\% \\
% %boolq & 52.02\% & 58.81\% & 62.20\% & 51.53\% & 40.09\% & 56.94\% \\
% %HellaSwag & 23.86\% & 23.28\% & 24.54\% & 23.11\% & 23.09\% & 23.55\% \\
% Obqa & 24.20\% & 24.40\% & 24.20\% & 24.00\% & 23.80\% & 19.80\% \\
% WinoGrande & 49.57\% & 51.70\% & 51.85\% & 50.28\% & 50.20\% & 48.30\% \\
% \bottomrule
% \end{tabular}
% \caption{Zero-Shot Accuracy (\%, $\uparrow$) Benchmarks for GPT-2 Models: Baseline, Midpoint, and Leapfrog.}
% \label{tab:gpt2_results_full_final}
% \end{table*}

\begin{table*}[t]
\centering
\footnotesize
\setlength{\tabcolsep}{4pt} % tighter columns
\begin{tabular}{@{}l ccc ccc@{}}
\toprule
& \multicolumn{3}{c}{\shortstack{\textbf{GPT-2 Small}\\ \textbf{(124M)}}}
& \multicolumn{3}{c}{\shortstack{\textbf{GPT-2 Large}\\ \textbf{(772M)}}} \\
\cmidrule(lr){2-4} \cmidrule(lr){5-7}
\textbf{Benchmark} & \textbf{Baseline} & \textbf{Midpoint} & \textbf{Leapfrog}
                   & \textbf{Baseline} & \textbf{Midpoint} & \textbf{Leapfrog} \\
\midrule
piqa        & 51.14\% & 53.05\% & 52.83\% & 53.16\% & 52.72\% & 55.11\% \\
ARC-e       & 27.02\% & 26.14\% & 28.95\% & 23.86\% & 26.84\% & 26.49\% \\
ARC-c       & 22.74\% & 22.73\% & 24.08\% & 22.41\% & 28.09\% & 26.09\% \\
Obqa        & 24.20\% & 24.40\% & 24.20\% & 24.00\% & 23.80\% & 19.80\% \\
WinoGrande  & 49.57\% & 51.70\% & 51.85\% & 50.28\% & 50.20\% & 48.30\% \\
\bottomrule
\end{tabular}
\caption{Zero-shot accuracy (\%, $\uparrow$) for GPT-2 models under baseline, midpoint, and leapfrog architectures.}
\label{tab:gpt2_results_full_final}
\end{table*}

\begin{figure}[t]
\centering

\subfloat[Training loss\label{fig:gpt2_train_losses}]{%
    \includegraphics[width=0.45\linewidth]{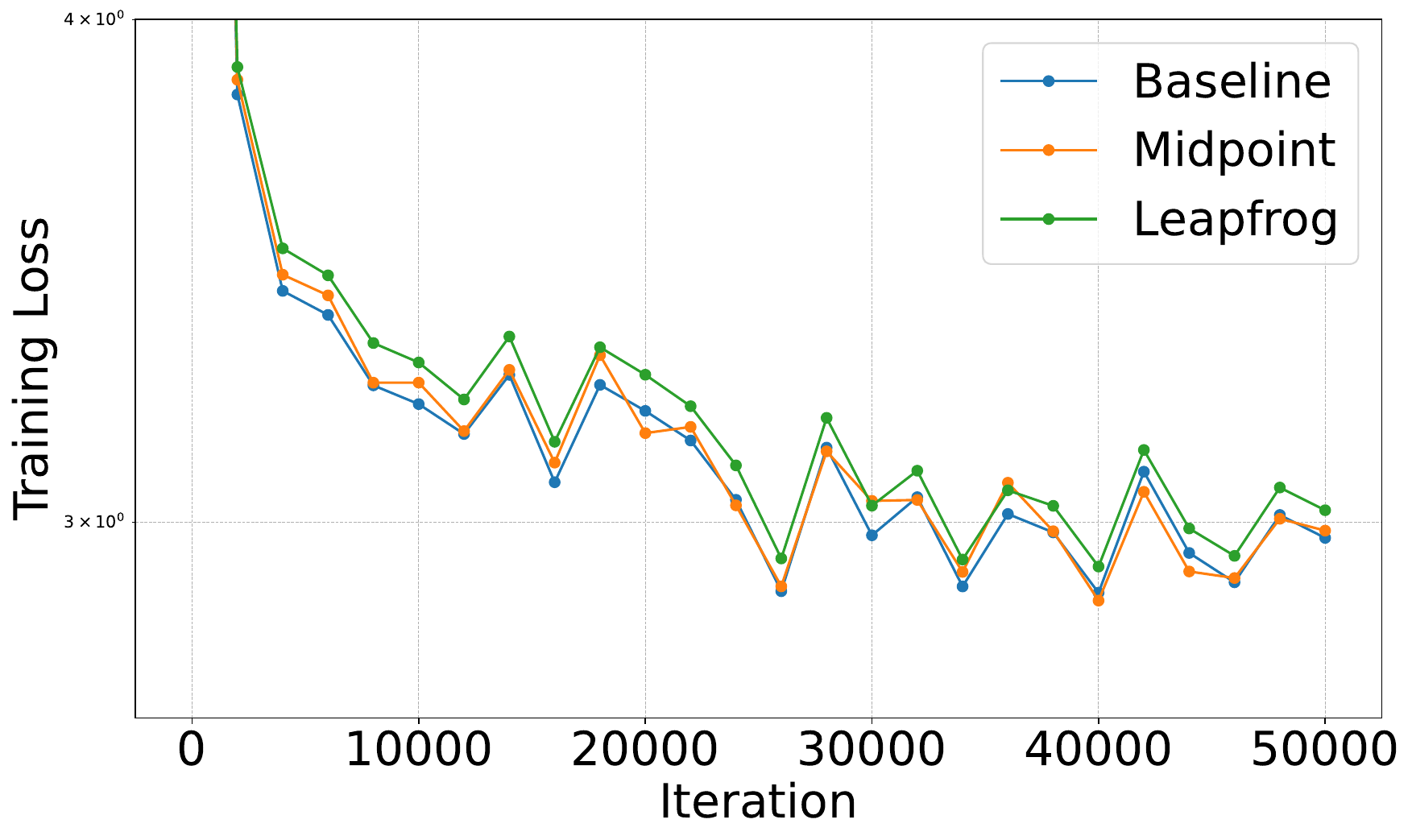}
}
\hfill
\subfloat[Validation loss\label{fig:gpt2_val_losses}]{%
    \includegraphics[width=0.45\linewidth]{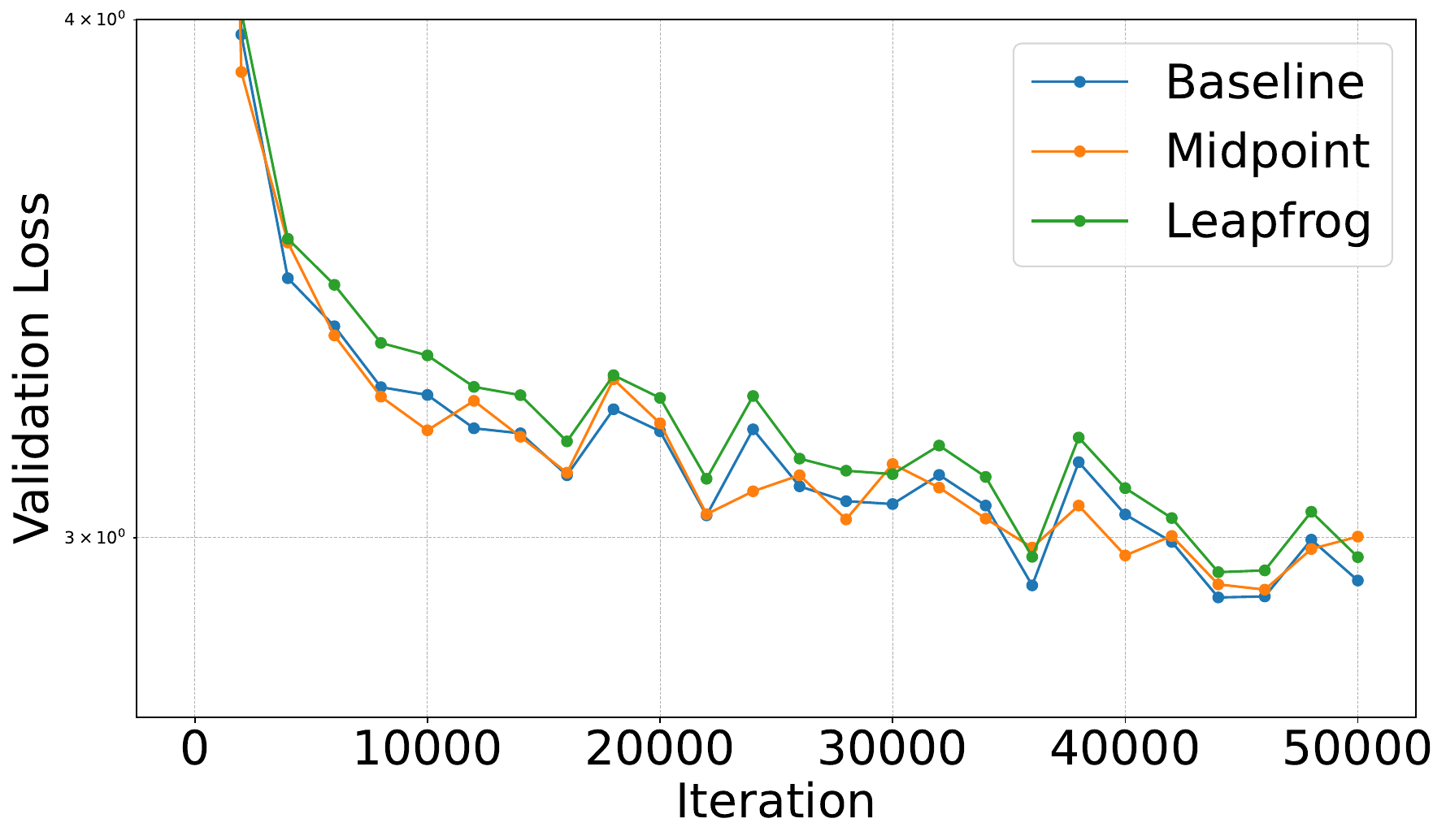}
}

\caption{Training and validation loss curves of GPT-2 using baseline, midpoint, and leapfrog architectures.}
\label{fig:gpt2_losses}
\end{figure}

% \begin{figure}[t]
%     \centering
%     \begin{subfigure}{0.5\textwidth}
%         \includegraphics[width=0.99\linewidth]{figures/gpt2_small_training_curve.pdf}
%         \caption{Training loss}
%         \label{fig:gpt2_train_losses}
%     \end{subfigure}\hfill
%     \begin{subfigure}{0.5\textwidth}
%         \includegraphics[width=0.99\linewidth]{figures/gpt2_small_validation_curve.pdf}
%         \caption{Validation loss}
%         \label{fig:gpt2_val_losses}
%     \end{subfigure}
%     \caption{Training and validation loss curves of GPT2 using baseline, midpoint, and leapfrog architectures.}
%     \label{fig:gpt2_losses}
% \end{figure}

\iffalse
Demonstrating that a particular architecture can outperform the most advanced one requires massive resources and massive data. Here, instead, we use rather modest resources and explore the following two research questions.
\begin{enumerate}
    \item Given the same number of parameters and the same compute budget, does a reversible architecture achieve comparable performance while using significantly less memory?
    \item Given a pre-trained model, can we convert it to a reversible model using only fine tuning? And if so, how does the model perform against some known benchmarks?
\end{enumerate}
Next, we explore the answers to these questions by conducting experiments on GPT2 architectures (small and large), as well as TinyLlama \cite{zhang2024tinyllama} and SmolLM2 \cite{allal2025smollm2smolgoesbig}  that both contain over a billion parameters.
\fi

Demonstrating that a new LLM architecture outperforms state-of-the-art models typically requires substantial computational and data resources. Here, we use modest resources to explore the following two research questions:
\begin{enumerate}
\item Given the same number of parameters and compute budget, can a reversible architecture achieve comparable performance while using significantly less memory?
\item Given a pre-trained model, can it be converted into a reversible one using only fine-tuning, and how does the resulting model perform on standard benchmarks?
\end{enumerate}

We address these research questions through experiments and evaluations on small and large GPT-2 \cite{radford2019language} architectures, as well as TinyLlama \cite{zhang2024tinyllama} and SmolLM2 \cite{allal2025smollm2smolgoesbig}, both of which contain over a billion parameters.

\subsection{Training a Reversible Model}

%We have trained two different reversible architectures, namely the Midpoint \Eqref{revnetforS} and Leapfrog \Eqref{eq:leapfrog}, on the OpenWebText dataset and compare them to the standard baseline residual skip-connection training. We chose to train Midpoint according to \Eqref{revnetforS} as it generalizes, by design, \Eqref{eq:leapfrog}, and we provide an ablation using \Eqref{eq:leapfrog} in the Appendix. Our proposed invertible methods are evaluated against the original GPT2 network as a baseline, with results presented in Figure~\ref{fig:gpt2_losses}. In this set of experiments, we use the hyperparameters and training procedure so that we can compare the convergence of the method on equal footings. The experiments demonstrate that our methods achieve comparable and even slightly better performance while preserving the desirable property of invertability.
We train two reversible architectures: Midpoint \eqref{revnetforS} and Leapfrog \eqref{eq:leapfrog}, on the OpenWebText dataset, and evaluate their performance against the standard residual skip-connection denoted Baseline, using both GPT-2 Small and GPT-2 Large (740M parameters). We primarily focus on the Midpoint with $\theta$ formulation, as it generalizes the Midpoint method from \eqref{eq:midpoint}. An ablation with the Midpoint method is provided in the Appendix. All models are trained using identical hyperparameters and procedures to ensure a fair comparison. Results for both model sizes are shown in Figure~\ref{fig:gpt2_losses} and summarized in Table~\ref{tab:gpt2_train_val}. As seen in the training curves and final validation metrics, the reversible architectures behave similarly to their non-reversible counterparts during training and exhibit slightly better validation loss. This demonstrates that reversible models not only are memory efficient, but also match or outperform standard residual models, without added training complexity.

%Similarly, we train the large GPT2 model with over 740M parameters and obtain similar results and training and validation. The training and validation summary of these models is presented in Table~\ref{tab:gpt2_train_val}. As can be seen by the curves as well as the final results, the training and validation of reversible architectures behave similarly to the non-reversible ones. Moreover, the final validation loss of the reversible architectures is slightly smaller than their corresponding non-reversible counterparts. This demonstrates that training a reversible architecture is not more complex than training a non-reversible one.  

\paragraph{Evaluation.}
%Although our reversible models achieved similar or slightly better validation when training on the OpenWebText dataset, we have evaluated their performance over 5 different datasets (zero-shot evaluation), namely piqa, ARC-e, ARC-c, Obqa, and WinoGrande. The results are presented in Table~\ref{tab:gpt2_results_full_final}.

%We note that reversible models are almost at par with the standard skip-connection models and in some cases outperform them. Thus, we conclude that for the architectures tested here, not only that reversible architectures are more memory efficient, they can also outperform standard architectures in terms of performance. 

To evaluate the generalization performance of our reversible Midpoint and Leapfrog models beyond the training distribution, we conduct zero-shot evaluation on five downstream benchmarks: piqa, ARC-easy, ARC-challenge, Obqa, and WinoGrande. These experiments are performed using models trained on the OpenWebText dataset, as described in Section~\ref{sec:conversion}. Results are reported in Table~\ref{tab:gpt2_results_full_final}.
Across tasks, the reversible architectures perform on par with their baseline, standard residual counterparts and, in several cases, outperform them. These findings suggest that reversible models not only provide substantial memory savings during training but also maintain competitive generalization performance without added training complexity.

\begin{figure}[t]
    \centering
    \includegraphics[width=0.7\linewidth]{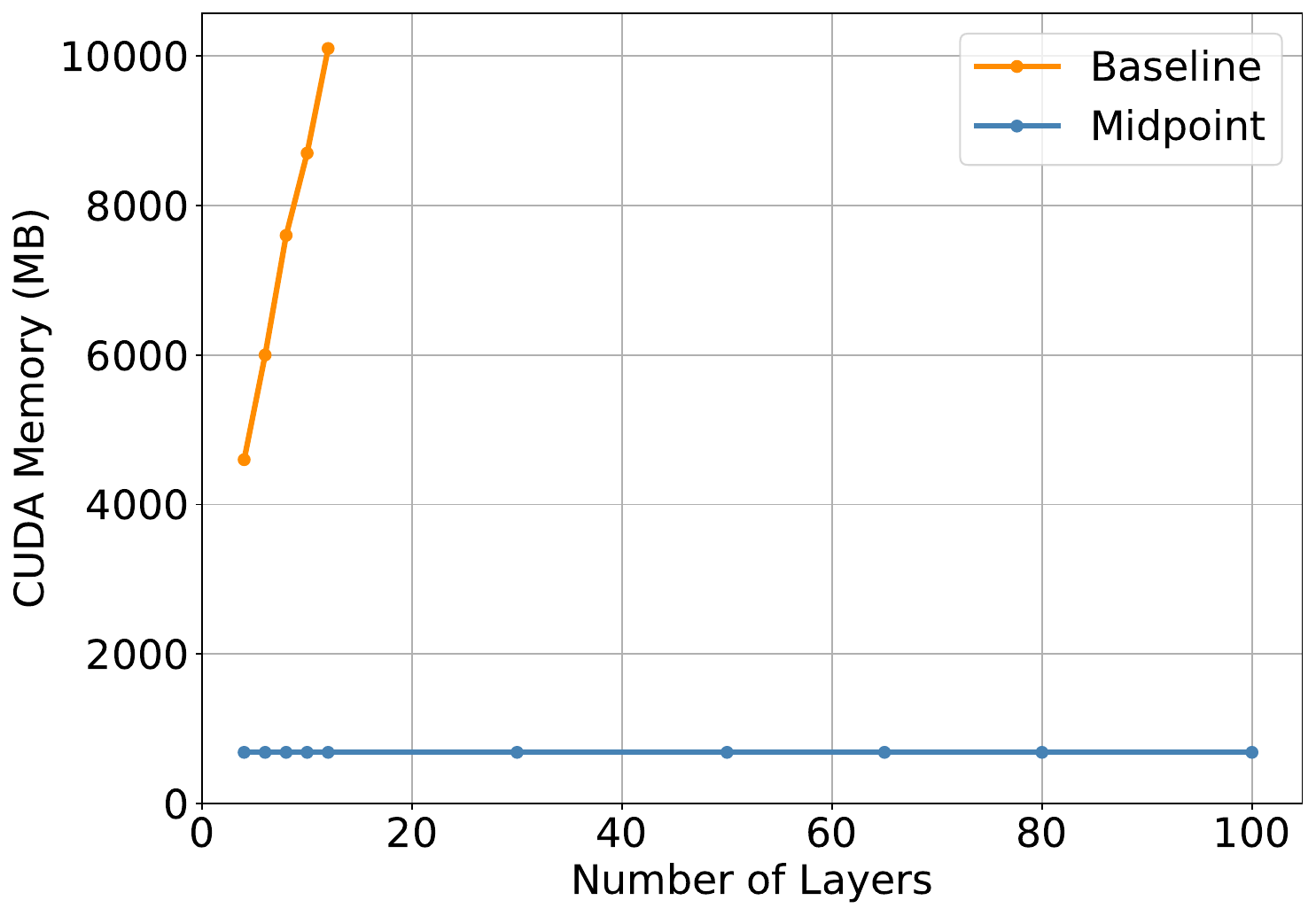}
    \caption{Training GPU memory usage vs. network depth for baseline and reversible (Midpoint) models. The baseline memory grows linearly and fails beyond 12 layers, while the reversible model remains constant.}
    \label{fig:memory_consumption}
\end{figure}

\paragraph{Memory Consumption.} While reversible architectures may yield modest improvements in accuracy, their key advantage lies in significantly reduced memory requirements during training. Table~\ref{tab:memory} reports the maximum batch size that fits in GPU memory for both baseline and reversible (Midpoint) models across a range of GPU architectures. The reversible architecture enables batch sizes nearly an order of magnitude larger than the baseline, specifically around 10$\times$ on all tested hardware. This highlights its practical benefit for memory-constrained training settings, enabling more efficient hardware utilization with no degradation in model quality. To further illustrate this advantage, Figure~\ref{fig:memory_consumption} shows memory usage as a function of model depth. While the memory footprint of standard residual networks grows linearly with depth, that of reversible models remains constant.

In Table~\ref{tab:batch_timing}, we report the throughput improvements achieved using reversible architectures. For a model with a hidden dimension of 512 and 8 attention heads, we observe that the benefit of our reversible updates increases with depth. In particular, deeper models support substantially larger batch sizes, leading to significant throughput gains, reaching up to 101\% improvement for the 96-layer model, and thereby enabling faster and more memory-efficient training.

\begin{table*}[t]
\centering
\begin{tabular}{llcccccc}
\toprule
Method & GPU &  RTX6000 & A10 & A100 & A6000 & H100 \\
 & Memory &  24GB & 24GB  & 40GB  & 48GB & 80GB \\
\midrule
Baseline & Max batch           & 6  & 6 & 12 &  15  & 26\\
Midpoint & Max batch   & 58 & 58 & 119 & 140 & 257  \\
\midrule
 & Enhancement &  9.66 & 9.66 & 9.91 & 9.33 & 9.88 \\
\bottomrule
\end{tabular}
\caption{Maximum batch size that fits in memory for baseline and reversible (Midpoint) networks across various GPU architectures. Enhancement indicates the relative improvement in batch size made possible by the reversible formulation.}
\label{tab:memory}
\end{table*}

\begin{table}[t]
\centering
\begin{tabular}{l l *{4}{c}}
\toprule
\textbf{Method} &  & \textbf{$L = 16$} & \textbf{$L = 32$} & \textbf{$L = 64$} & \textbf{$L = 96$} \\
\midrule
\multirow{3}{*}{Baseline}
  & Batch Size    & 326 & 176 & 88 & 52 \\
  & Time (s)      & 0.667 & 0.669 & 0.774 & 0.913 \\
  & Throughput    & 488.8 & 263.1 & 113.7 & 56.96 \\
\midrule

\multirow{3}{*}{Midpoint}
  & Batch Size    & 1550 & 1374 & 1076 & 874 \\
  & Time (s)      & 2.627 & 4.207 & 6.362 & 7.634 \\
  & Throughput    & 590.0 & 326.6 & 169.1 & 114.49 \\
\midrule

\multirow{1}{*}{Throughput Gain (\%)} 
  &               & 20.7\% & 24.1\% & 48.8\% & 101.0\% \\
\bottomrule
\end{tabular}
\caption{Maximum batch size, per-step training time (s), and resulting throughput (samples/s) for different model depths, as a function of number of layers $L$.}
\label{tab:batch_timing}
\end{table}

\subsection{Converting Baseline to Reversible LLMs}

\begin{figure}[t]
    \centering        \includegraphics[width=0.7\linewidth]{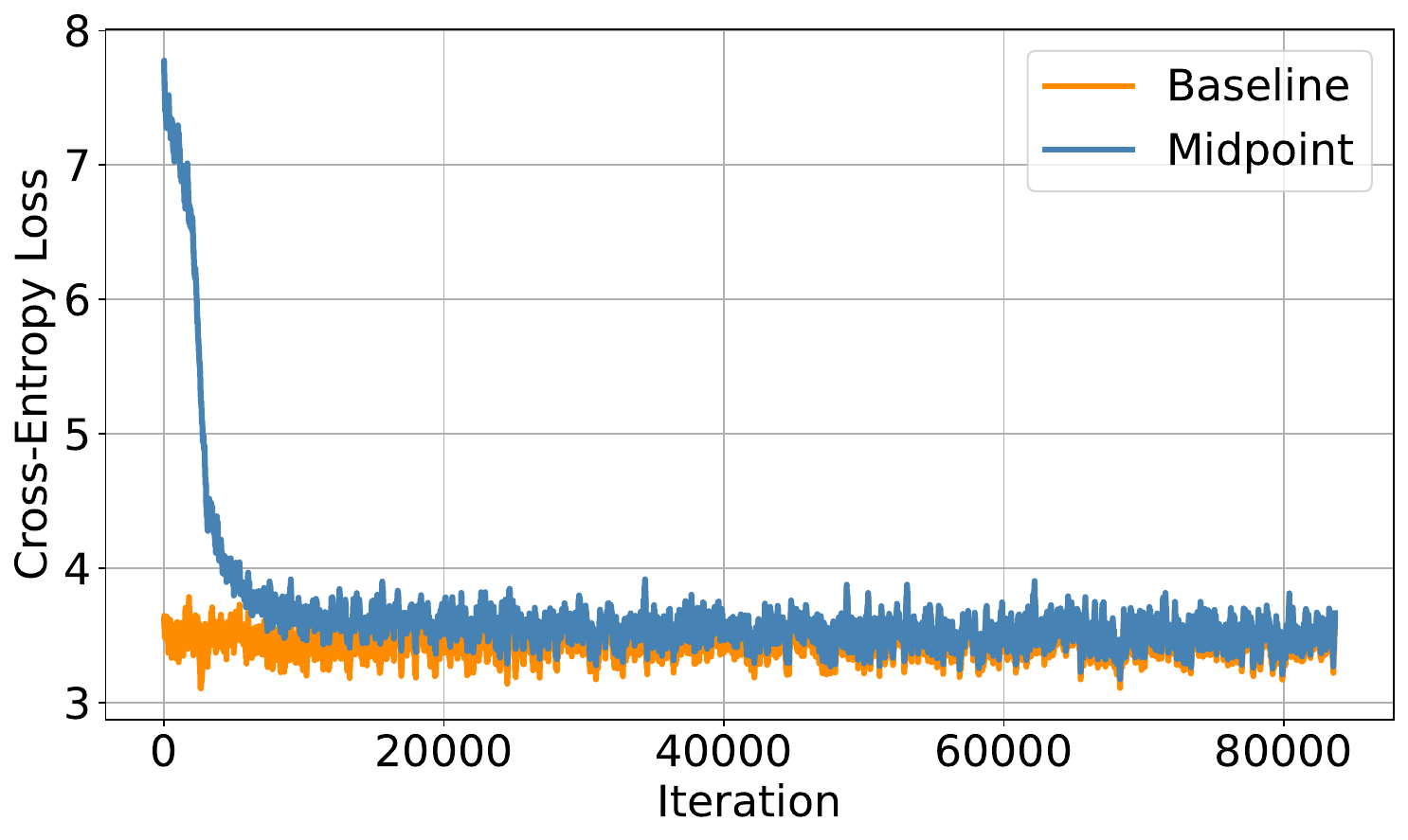}
        \caption{Next token prediction cross-entropy loss during the conversion of TinyLlama-v1.0 to a reversible Midpoint architecture. The reversible model (Midpoint) closely matches the baseline in next-token prediction, demonstrating successful functional alignment.}
        \label{fig:train_conversion}
\end{figure}

%While large benefits can be obtained by training a reversible method, it can be highly beneficial to obtain a reversible network even when the original network was trained in a non-reversible way. To do that, we conduct experiments using TinyLlama, where we take a pre-trained model and use the dynamics defined by  \Eqref{eq:almostmid_rev} to convert the non-reversible model into a reversible one.To this end, we use the WikiText data set to match the trajectories of the reversible and the non-reversible architectures, as presented in \Eqref{min_opt1}. We run only 80,000 batches to obtain the conversion and observe a good fit on the validation set.
%The training and validation obtained in the process are presented in Figure~\ref{fig:train_conversion}.
While training a reversible model from scratch offers substantial advantages, retrofitting a pre-trained \allowbreak\ non-reversible model can be equally valuable. To demonstrate this, we convert TinyLlama-v1.0 using the reversible dynamics in \eqref{eq:almostmid_rev}. We align the output distributions of the original and converted models on the WikiText dataset via the objective in \eqref{min_opt1}. The conversion uses only 80,000 examples and yields a close match on the validation set. As shown in Figure~\ref{fig:train_conversion}, the reversible model closely follows the training and validation loss of the original in next-token prediction. These results confirm that the conversion preserves model behavior and quality while enabling reversibility.

Since the reversible model is designed to approximate the behavior of its non-reversible counterpart, we expect its performance to closely match, but not necessarily exceed, that of the original model. To evaluate the generalization capability of the converted reversible network, we assess its performance on a suite of zero-shot commonsense benchmarks. Results are reported in Table~\ref{tab:tiny_llama_zeroshot_commonsense}.
\textit{Additionally}, we include results for a converted SmolLM2 model with 1.7B parameters on the MMLU benchmark, provided in the Appendix. This model was retrofitted using two epochs of WikiText data and achieves an accuracy of 49\%, closely matching the original model’s 50.36\%. Overall, the converted and original models exhibit highly similar performance. These results demonstrate that reversible architectures can be effectively obtained from pre-trained models and are well-suited for fine-tuning with minimal overhead.

% \begin{table*}[h]
% \centering
% \begin{tabular}{lcccccc}
% \toprule
% Model & Obqa & WinoGrande & ARC-c & ARC-e & boolq & piqa  \\
% \midrule
% TinyLlama v1.0 & 38.20 & 50.36 & 27.76 & 50.88 & 60.92 & 72.09  \\
% Reversible Model & 35.60 & 51.62 & 26.75 & 50.35 & 60.83 & 70.24 \\  
% \bottomrule
% \end{tabular}
% \caption{Conversion (non-reversible to reversible) performance on commonsense reasoning tasks (Reversible obtained with $\theta$)}
% \label{tab:tiny_llama_zeroshot_commonsense}
% \end{table*}

\iffalse
\begin{table*}[t]
\centering
\begin{tabular}{lccccccc}
\toprule
Model & HellaSwag & Obqa & WinoGrande & ARC-c & ARC-e & boolq & piqa  \\
\midrule
Baseline & 58.74 & 38.20 & 50.36 & 27.76 & 50.88 & 60.92 & 72.09  \\
Midpoint & 49.89 & 35.60 & 51.62 & 26.75 & 50.35 & 60.83 & 70.24 \\  
% Semi-Reversable & 39.57 & 34.00 &  & 27.42 & 51.23  & 63.67 & 70.24 &  \\
\bottomrule
\end{tabular}
\caption{Performance on commonsense reasoning tasks on a baseline TinyLlamav1.0 and its conversion into a  Midpoint TinyLlamav1.0 architecture.}
\label{tab:tiny_llama_zeroshot_commonsense}
\end{table*}
\fi

\begin{table}[t]
\centering
\begin{tabular}{lcc}
\toprule
\textbf{Benchmark} & \textbf{Baseline} & \textbf{Baseline $\rightarrow$ Midpoint} \\
\midrule
piqa & 72.09\% & 70.24\% \\
ARC-e & 50.88\% & 50.35\% \\
ARC-c & 27.76\% & 26.75\% \\
Obqa & 38.20\% & 35.60\% \\
WinoGrande & 50.36\% & 51.62\% \\
\bottomrule
\end{tabular}
\caption{Zero-shot accuracy (\%, $\uparrow$) on commonsense reasoning benchmarks. We compare the original TinyLlama-v1.0 model to its reversible counterpart obtained via conversion with the Midpoint method. The reversible model maintains comparable performance across all tasks.}
\label{tab:tiny_llama_zeroshot_commonsense}
\end{table}

\subsection{Ablation Study}
\label{app:ablation_study}
We conduct an ablation study comparing the midpoint methods described in  \eqref{eq:midpoint} and \eqref{revnetforS}, and show our results in \Cref{tab:gpt2_results_updated}. While the results are often comparable, the method introduced in  \eqref{revnetforS} may demonstrate potential for significant improvement.

% \begin{table*}[h]
% \centering
% \begin{tabular}{l ccc cc}
% \toprule
% & \multicolumn{3}{c}{\textbf{GPT-2 Small (123.59M)}} & \multicolumn{2}{c}{\textbf{GPT-2 Large (772.27M)}} \\
% \cmidrule(lr){2-4} \cmidrule(lr){5-6}
% \textbf{Benchmark} & \textbf{Midpoint (no $\theta$)} & \textbf{Midpoint ($\theta$)} & \textbf{Baseline} & \textbf{Midpoint ($\theta$)} & \textbf{Baseline} \\
% \midrule
% piqa & 50.98\% & 53.05\% & 51.14\% & 52.72\% & 53.16\% \\
% ARC-e & 26.49\% & 26.14\% & 27.02\% & 26.84\% & 23.86\% \\
% ARC-c & 23.08\% & 22.73\% & 22.74\% & 28.09\% & 22.41\% \\
% boolq & 59.85\% & 58.81\% & 52.02\% & 40.09\% & 51.53\% \\
% HellaSwag & 23.42\% & 23.28\% & 23.86\% & 23.09\% & 23.11\% \\
% Obqa & 20.60\% & 24.40\% & 24.20\% & 23.80\% & 24.00\% \\
% WinoGrande & 50.75\% & 51.70\% & 49.57\%  & 50.20\% & 50.28\% \\
% \bottomrule
% \end{tabular}
% \caption{Zero-Shot Accuracy (\%) on Various Benchmarks for GPT-2 Models.}
% \label{tab:gpt2_results_updated}
% \end{table*}

\begin{table*}[h]
\centering
\begin{tabular}{l ccc}
\toprule
& \multicolumn{3}{c}{\textbf{GPT-2 Small (123.59M)}} \\
\cmidrule(lr){2-4}
\textbf{Benchmark} & \textbf{Midpoint ( \eqref{eq:midpoint})} & \textbf{Midpoint (  \eqref{revnetforS})} & \textbf{Baseline} \\
\midrule
piqa & 50.98\% & 53.05\% & 51.14\% \\
ARC-e & 26.49\% & 26.14\% & 27.02\% \\
ARC-c & 23.08\% & 22.73\% & 22.74\% \\
% boolq & 59.85\% & 58.81\% & 52.02\% \\
% HellaSwag & 23.42\% & 23.28\% & 23.86\% \\
Obqa & 20.60\% & 24.40\% & 24.20\% \\
WinoGrande & 50.75\% & 51.70\% & 49.57\% \\
\bottomrule
\end{tabular}
\caption{Zero-Shot Accuracy (\%) on Various Benchmarks for GPT-2 Small Model.}
\label{tab:gpt2_results_updated}
\end{table*}

\subsection{Experimental Details}
% \label{app:experimental_details}
% For training the GPT-2 models, we utilized 8 NVIDIA L40S GPUs. Model conversion was performed on an RTX6000 GPU. We adopted standard hyperparameters following the configurations provided in  https://github.com/karpathy/nanoGPT. During conversion, we used a learning rate of $10^{-5}$.
\label{app:experimental_details}
For training the GPT-2 models, we utilized 8 NVIDIA L40S GPUs. Model conversion was performed on an RTX6000 GPU. We adopted standard hyperparameters following the configurations provided in \url{https://github.com/karpathy/nanoGPT}. During conversion, we used a learning rate of $10^{-5}$.

\subsection{Additional Results: MMLU}
\label{app:additional_results}
The MMLU (Massive Multitask Language Understanding) dataset is a comprehensive benchmark for assessing language models’ reasoning and domain knowledge across a wide range of tasks. Our results, shown in Table \ref{tab:mmlu_topics_smollm2}  demonstrate that the proposed methods achieve performance comparable to, and in some cases exceeding, that of standard approaches.

{\small
% was p{4.2cm}
% \begin{longtable}{p{2.6cm} r r r r r}
\begin{longtable}{P{2.6cm} r r r r r}
\caption{Full MMLU subject-level accuracy for SmolLM2 1.7B before and after reversible conversion.}
\label{tab:mmlu_topics_smollm2} \\
\toprule
\textbf{Subject} 
& \textbf{Samples} 
& \textbf{\shortstack{Original \\ Acc}} 
& \textbf{\shortstack{Original \\ Std. Error}} 
& \textbf{\shortstack{Converted \\ Acc}} 
& \textbf{\shortstack{Converted \\ Std. Error}} \\
\midrule
\endfirsthead

\toprule
\textbf{Subject} 
& \textbf{Samples} 
& \textbf{\shortstack{Original \\ Acc}} 
& \textbf{\shortstack{Original \\ Std. Error}} 
& \textbf{\shortstack{Converted \\ Acc}} 
& \textbf{\shortstack{Converted \\ Std. Error}} \\
\midrule
\endhead

all & 14042 & 0.5036 & 0.0042 & 0.4900 & 0.0042 \\
\midrule
abstract algebra              & 100& 0.2900 & 0.0454 & 0.3100 & 0.0462 \\
anatomy                       & 135& 0.4370 & 0.0427 & 0.4148 & 0.0424 \\
astronomy                     & 152& 0.6118 & 0.0395 & 0.5987 & 0.0398 \\
business ethics               & 100& 0.6000 & 0.0490 & 0.5600 & 0.0496 \\
clinical knowledge            & 265& 0.5660 & 0.0304 & 0.5509 & 0.0306 \\
college biology               & 144& 0.5556 & 0.0414 & 0.5208 & 0.0416 \\
college chemistry             & 100& 0.3800 & 0.0485 & 0.3500 & 0.0477 \\
college computer science      & 100& 0.4500 & 0.0497 & 0.4300 & 0.0495 \\
college mathematics           & 100& 0.3400 & 0.0474 & 0.3800 & 0.0485 \\
college medicine              & 173& 0.4971 & 0.0380 & 0.4509 & 0.0378 \\
college physics               & 102& 0.3431 & 0.0470 & 0.3431 & 0.0470 \\
computer security             & 100& 0.6200 & 0.0485 & 0.6200 & 0.0485 \\
conceptual physics            & 235& 0.4809 & 0.0326 & 0.4638 & 0.0325 \\
econometrics                  & 114& 0.3596 & 0.0449 & 0.3947 & 0.0458 \\
electrical engineering        & 145& 0.5241 & 0.0415 & 0.5379 & 0.0414 \\
elementary mathematics        & 378& 0.3783 & 0.0249 & 0.3889 & 0.0251 \\
formal logic                  & 126& 0.3254 & 0.0417 & 0.3175 & 0.0415 \\
global facts                  & 100& 0.3100 & 0.0462 & 0.3000 & 0.0458 \\
high school biology           & 310& 0.5323 & 0.0283 & 0.5516 & 0.0282 \\
high school chemistry         & 203& 0.4631 & 0.0350 & 0.4286 & 0.0347 \\
high school computer science  & 100& 0.5000 & 0.0500 & 0.4600 & 0.0498 \\
high school european history  & 165& 0.5636 & 0.0386 & 0.5636 & 0.0386 \\
high school geography         & 198& 0.6162 & 0.0346 & 0.5909 & 0.0349 \\
high school government and politics & 193& 0.7254 & 0.0321 & 0.7098 & 0.0327 \\
high school macroeconomics    & 390& 0.4923 & 0.0253 & 0.4615 & 0.0252 \\
high school mathematics       & 270& 0.3296 & 0.0286 & 0.3222 & 0.0284 \\
high school microeconomics    & 238& 0.5042 & 0.0324 & 0.4622 & 0.0323 \\
high school physics           & 151& 0.3576 & 0.0390 & 0.3444 & 0.0387 \\
high school psychology        & 545& 0.6936 & 0.0197 & 0.6514 & 0.0204 \\
high school statistics        & 216& 0.4074 & 0.0334 & 0.3611 & 0.0327 \\
high school US history        & 204& 0.5588 & 0.0348 & 0.5637 & 0.0347 \\
high school world history     & 237& 0.6076 & 0.0317 & 0.5823 & 0.0320 \\
human aging                   & 223& 0.6054 & 0.0327 & 0.6009 & 0.0328 \\
human sexuality               & 131& 0.6031 & 0.0427 & 0.6031 & 0.0427 \\
international law             & 121& 0.6529 & 0.0433 & 0.6529 & 0.0433 \\
jurisprudence                 & 108& 0.5833 & 0.0474 & 0.5833 & 0.0474 \\
logical fallacies             & 163& 0.6503 & 0.0374 & 0.6380 & 0.0376 \\
machine learning              & 112& 0.3482 & 0.0450 & 0.3214 & 0.0441 \\
management                    & 103& 0.7087 & 0.0448 & 0.6893 & 0.0456 \\
marketing                     & 234& 0.7222 & 0.0293 & 0.7222 & 0.0293 \\
medical genetics              & 100& 0.5900 & 0.0492 & 0.5700 & 0.0495 \\
miscellaneous                 & 783& 0.6986 & 0.0164 & 0.6858 & 0.0166 \\
moral disputes                & 346& 0.5520 & 0.0267 & 0.5376 & 0.0268 \\
moral scenarios               & 895& 0.2380 & 0.0142 & 0.2369 & 0.0142 \\
nutrition                     & 306& 0.5523 & 0.0284 & 0.5523 & 0.0284 \\
philosophy                    & 311& 0.6141 & 0.0276 & 0.6431 & 0.0272 \\
prehistory                    & 324& 0.5957 & 0.0273 & 0.5895 & 0.0273 \\
professional accounting       & 282& 0.3369 & 0.0281 & 0.3262 & 0.0279 \\
professional law              & 1534& 0.3683 & 0.0123 & 0.3449 & 0.0121 \\
professional medicine         & 272& 0.4118 & 0.0298 & 0.3934 & 0.0296 \\
professional psychology       & 612& 0.5245 & 0.0202 & 0.4951 & 0.0202 \\
public relations              & 110& 0.5091 & 0.0477 & 0.5636 & 0.0473 \\
security studies              & 245& 0.5878 & 0.0314 & 0.5755 & 0.0316 \\
sociology                     & 201& 0.6667 & 0.0333 & 0.6318 & 0.0340 \\
US foreign policy             & 100& 0.7500 & 0.0433 & 0.7700 & 0.0421 \\
virology                      & 166& 0.4880 & 0.0388 & 0.4639 & 0.0387 \\
world religions               & 171& 0.7193 & 0.0344 & 0.6901 & 0.0354 \\
\bottomrule
\end{longtable}
}

\section{Summary and Outlook}

We presented a new class of \emph{reversible large language models} derived from first and second order dynamics.  
Conventional transformers must cache every intermediate activation for back-propagation, creating a severe memory bottleneck.  
Our Leapfrog, Midpoint, and Hamiltonian updates are \emph{reversible by construction}; hidden states are reconstructed during the backward pass instead of stored, yielding up to a 10$\times$ reduction in activation memory and enabling much larger batch sizes.

\textbf{Numerical PDE Methods in the Machine Learning Context.} It is important to keep in mind that our employment of numerical methods for PDEs does not mean that we are interested in numerically solving PDE systems in this paper. Specifically, numerical methods that may fail in the PDE context (especially when a highly accurate numerical solution is required) may not necessarily be inadequate in our present context. Noise can have a therapeutic effect!

The clearest example for this effect is the explicit midpoint method \ref{eq:midpoint}. As a PDE method it is not even compact, as it spans over two mesh intervals in ``time’’ while related to a 1st order PDE. Thus, fast waves of the form

\[
p^{\ell} =
\begin{cases}
m, & \text{if $\ell$ is odd}, \\
n, & \text{if $\ell$ is even}.
\end{cases}
\]

will produce zero in the left hand side of \ref{eq:midpoint} for any real values $m$ and $n$.
However, in the ML domain (unlike for predicting a tsunami) this method can be successful, which is a side effect of dealing with many noisy observations.

In contrast, the methods presented in \ref{eq:leapfrog} and \ref{eq:hamil}, which approximate the 2nd order PDE (\ref{eq:second-order}), are compact. And yet, they too could suffer from a catastrophic accumulation of roundoff errors when applied in the traditional numerical mathematics context of approximating an exact solution for a nonlinear PDE system over a long time interval; see for instance \cite{faou12,brlp,asma1}. However, again in the present context, delicate usage of symmetric methods, which occasionally gives the nod to symplectic methods in the context of accurately solving models such as Korteweg de Vries and nonlinear Schrodinger, does not buy much leverage here.  

\textbf{Theoretical and empirical contributions.}  
We provide a stability analysis that pinpoints conditions for forward- and backward-stable training.  
On GPT-2 (124M/772M), TinyLlama (1.1B), and SmolLM2 (1.7B), R-LLMs match or slightly surpass the perplexity and zero-shot accuracy of non-reversible baselines on PIQA, ARC-easy/-challenge, OBQA, WinoGrande, and MMLU, demonstrating that memory efficiency need not compromise quality.  
In addition, we introduce a lightweight fine-tuning procedure that converts existing checkpoints into fully reversible models using only two epochs of auxiliary data, making our approach immediately applicable across today’s LLM ecosystem.

\textbf{Future directions.} We identify the following research paths:
\begin{enumerate}[leftmargin=*]
\item \textbf{Frontier models.}  Extend reversible updates and retrofit to hundred-billion-parameter foundation models. 
\item \textbf{Richer conservative dynamics.}  Explore variational, symplectic, or other \allowbreak\ Hamiltonian integrators as inductive biases. 
\item \textbf{Long-context training.}  Exploit constant-memory back-propagation to handle document-scale or code-base sequences. 
\item \textbf{Faster retrofit objectives.}  Design loss functions and parameter-freezing schemes that further cut conversion cost. 
\end{enumerate}

By decoupling memory from depth, reversible architectures offer a practical, scalable path toward the next generation of efficient and extensible language models.

\newpage

\bibliographystyle{siamplain}
\bibliography{references}

\newpage
\onecolumn

\end{document}